%% file: icra2027-prime.tex
\documentclass[letterpaper, 10 pt, conference]{ieeeconf}  

\IEEEoverridecommandlockouts                              

\usepackage{graphicx} 
\usepackage{subcaption}
\usepackage{amsmath} 
\usepackage{bm} 
\usepackage{url}
\usepackage{cite}
\usepackage{hyperref}
\usepackage[usenames,dvipsnames,table]{xcolor}

\usepackage{booktabs}
\usepackage{xspace}
\usepackage{multirow}
\usepackage{amsfonts}
\usepackage{diagbox}
\usepackage{makecell}
\usepackage[table]{xcolor}
\definecolor{lightgray}{rgb}{0.9,0.9,0.9}

\usepackage{etoolbox}

\renewcommand{\citedash}{\hbox{-}\penalty\citepunctpenalty}
\makeatletter
\patchcmd{\@compress@cite}
  {\expandafter\def\expandafter\@h@ld\expandafter{\@citea\@cite@out{#3}}}
  {\def\@h@ld{\citedash\@cite@out{#3}}}
  {}
  {\PackageError{citation-format}{Cannot enable two-number citation ranges}
    {Check the installed cite package before compiling.}}
\makeatother

\hypersetup{
    colorlinks=true,
    linkcolor=MidnightBlue,
    filecolor=magenta,      
    urlcolor=MidnightBlue,
    citecolor=black,
}
\usepackage[capitalise, nameinlink]{cleveref}

\newcommand{\name}{PRIME}

\newcommand\blfootnote[1]{%
\begingroup
\renewcommand\thefootnote{}\footnote{#1}%
  \addtocounter{footnote}{-1}%
  \endgroup
}

\begin{document}

\bstctlcite{IEEEauthorLimit}

\title{\LARGE \bf Not All Actions Contribute Equally: Learning Precise Progress Estimation from Imprecise Demonstrations}

\author{Huanqian Wang$^{1*}$, Boshi Tang$^{1*}$, Botong Xiao$^{1}$, Jie Zhang$^1$, Yang Yue$^1$,\\ Danhua Tao$^1$,  Shiji Song$^1$, Gao Huang$^{1\dagger\,}$}

\maketitle

\begin{abstract}

Progress estimators play an essential role in providing dense reward signals for reinforcement learning in robotic manipulation, but accurately assessing task progress can be much challenging.
Existing methods heavily rely on expert data and naively assume that all actions within a demonstration contribute to task success. 
This assumption often breaks down in real-world scenarios, especially in high-precision manipulation, where subtle geometric errors can lead to task failure.
Therefore, these methods can erroneously interpret human mistakes as meaningful progress, further corrupting progress learning.
We argue that a key limitation is the lack of a global prior regarding task evaluation criteria, which is crucial for distinguishing meaningful progress from human mistakes.
To this end, we propose \name{}, a novel progress estimator capable of learning precise progress estimation from imprecise demonstrations by aggregating experience across trajectories.
\name{} constructs a weighted graph over observed states, connecting states across trajectories based on spatial proximity and visual consistency. Edge weights encode spatial distances as transition costs, and shortest-path search approximates the geodesic distance to success. Changes in this distance naturally distinguish advancement, stagnation, and setbacks without assuming monotonic progress over time. 
Extensive experiments in both simulation and real world on precise manipulation tasks show that \name{} is consistently effective across diverse tasks, and robust to data with varying degrees of suboptimality.
Notably, the advantage of \name{} gets more pronounced in higher-precision scenarios, yielding up to a \(5\times\) improvement in policy success rate over the SFT baselines on charger insertion.

\end{abstract}

\blfootnote{\hspace{-2mm} $^*$Equal contribution. $^\dagger$Corresponding author. $^1$BNRist, Tsinghua University. Code: \url{https://github.com/lucywang720/PRIME}.}

\input{document/introduction}

\input{document/related}

\input{document/method}

\input{document/experiment}

\input{document/conclusion}

\begingroup
\hypersetup{urlcolor=black}
\bibliographystyle{IEEEtran}
\bibliography{references}
\endgroup

\clearpage
\appendix
\input{document/appendix}

\end{document}

%% file: document/introduction.tex
\section{Introduction}
\label{sec:intro}

Real-world Reinforcement Learning (RL) has emerged as an indispensable paradigm for advancing robotic manipulation from simple, repetitive actions to highly complex, precise skills~\cite{xu2026rl, li2025gr, huang2025dexterous, chen2026borabridgingofflinereinforcement}. 
However, mastering these high-precision tasks demands feedback that goes beyond binary success signals to identify and rectify the inevitable imperfections within human-collected demonstrations and policy rollouts~\cite{xu2026rl, li2025gr, chen2025rm}.
To bridge this gap, progress estimators have garnered extensive attention in the research community~\cite{zhang2025rewind, chen2025sarmstageawarerewardmodeling, liang2026robometerscalinggeneralpurposerobotic, lv2026vivavideogenerativevaluemodel, mavip}.
Despite recent progress, faithfully capturing subtle advancement, stagnation, and setbacks remains challenging for progress estimators, particularly in high-precision manipulation~\cite{nair2018overcoming, wu2025learning}.

\begin{figure}[t]
    \centering
        \includegraphics[width=\linewidth]{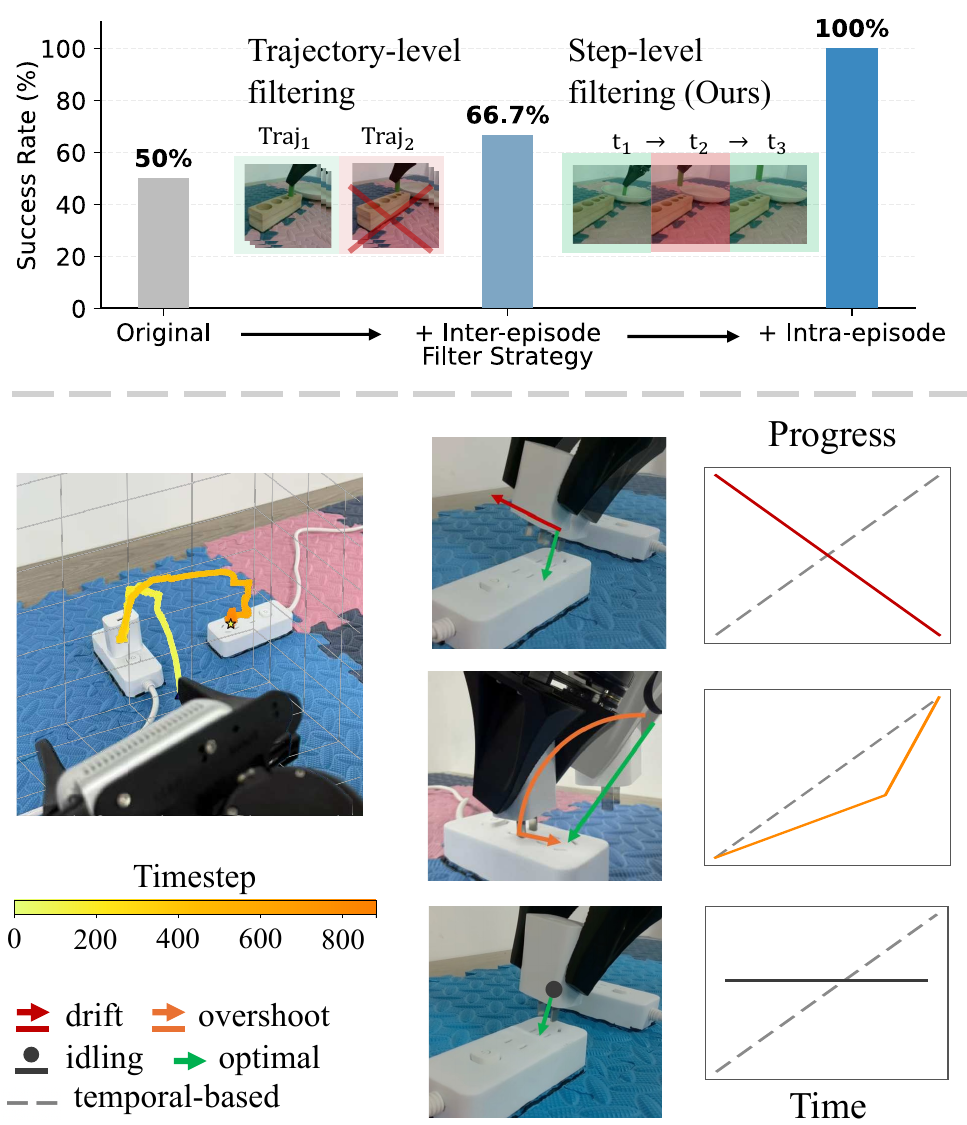}
    \caption{\textbf{Overview of suboptimal behaviors and their impact on policy training.} 
    \textbf{(Top)} Demonstrations for precision manipulation inevitably contain suboptimal behaviors, even after trajectory-level filtering. In peg-insertion, for example, removing successful yet low-quality trajectories improves the task success rate only from 50\% to 66.7\%, whereas our step-level filtering achieves a success rate of 100\%.
    \textbf{(Bottom)} During data collection, human operators unavoidably introduce suboptimal behaviors such as drifting (unintended lateral deviation), overshooting (motion beyond the target), and idling (unnecessary pauses or inactivity). While prior works assume that all actions contribute equally to the final success, they fail to capture the real-time progress in the face of these behaviors.}
    
    \label{teaser}
\end{figure}

\begin{figure*}[ht]
    \centering
        \includegraphics[width=\linewidth]{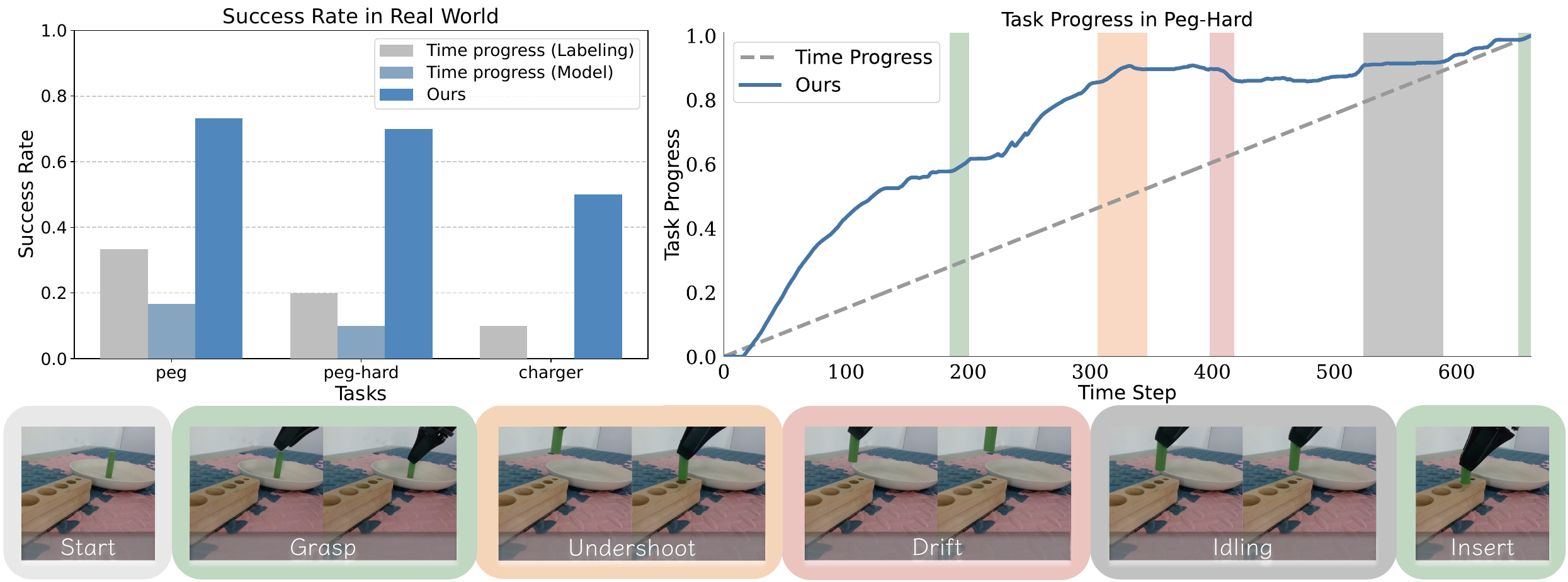} 
    \caption{\textbf{Quantitative success rate comparison and qualitative illustration of progress estimation.} \textbf{(Left)} Success rates of different methods across three real-world tasks: peg-in-hole, peg-in-hole (Hard) and charger insertion. \textbf{(Right)} Qualitative illustration of progress estimation on Peg-Hard task. \name{} reflects partial advancements during an undershoot, maintains a constant estimation during idling, and appropriately penalizes the progress when a drift occurs. \textbf{(Bottom)} Correspondence between suboptimal behaviors and progress estimation of PRIME. Our method accurately identifies different types of suboptimal behavior, and assigns progress estimation accordingly.}
    \label{intro-visual}
\end{figure*}

The inefficiency of existing methods in high-precision tasks stems from their inability to distinguish genuine task progress from suboptimal behaviors such as drifting, overshooting, and idling (Fig.~\ref{teaser}, Bottom). These methods rely on a simple yet flawed assumption that all actions within a trajectory contribute equally to eventual task success~\cite{liang2026robometerscalinggeneralpurposerobotic,lv2026vivavideogenerativevaluemodel}. Consequently, they depend heavily on expert data and discard a substantial proportion of collected demonstrations that fail to meet the required precision threshold. 
However, discarding low-quality demonstrations is costly and overlooks that, in high-precision settings, even high-quality demonstrations may contain steps that do not advance the task. Although trajectory-level filtering improves the policy success rate by 16.7 percentage points in our SFT experiments (Fig.~\ref{teaser}, top), suboptimal behaviors within the retained demonstrations can still confound progress estimates. This motivates finer-grained, step-level progress estimation.

Our key observation is that accurate progress estimation can be achieved only if the underlying progress structure captures holistic task priors. Such priors cannot be obtained from a single trajectory.
For example, when learning to operate an entirely new machine, the human operators cannot determine whether the current operation is optimal and can only attempt to faithfully follow the demonstration.
However, after observing multiple demonstrations, the operator can distinguish between suboptimal and optimal actions and thus estimate the progress of the current operation.
Thus, we incorporate this holistic information into our progress labels by leveraging the collective experience across multiple demonstrations, rather than relying solely on the temporal position within an individual trajectory.

Building on this observation, we propose PRIME (\textbf{PR}ecise progress estimation from \textbf{IM}precise d\textbf{E}monstrations), a progress estimation method extracting collective experience from the collected suboptimal demonstrations to assess step-level quality. We adopt a graph search framework to approximate geodesic distances over the feasible manifold.
Each node represents an observed state comprising the robot configuration and multi-view visual observations.
Edges are established based on configuration proximity and visual consistency, while edge weights quantify geometric changes in robot configuration, which allows for transitions to alternative trajectories with only a fine-grained distance penalty. Shortest-path search on the resulting graph estimates the geodesic distance to success. Changes in this distance naturally distinguish advancement, stagnation, and setbacks without assuming monotonic progress over time.

We validate the effectiveness of \name{} through extensive experiments in both real-world and simulation scenarios, which demonstrate that a policy trained with \name{} significantly improves the success rate across diverse tasks and is robust to data with varying degrees of suboptimality.
In real world experiments, \name{} achieves consistently superior success rates especially on higher-precision tasks, outperforming the standard SFT baseline by up to $2.2\times-5\times$ across 3 high-precision tasks: peg-in-hole (73.3\% v.s. 33.3\%), peg-in-hole Hard (70.0\% v.s. 20.0\%), and charger insertion (50.0\% v.s. 10.0\%).
We further construct a diagnostic benchmark covering stage understanding, local precision, and progress continuity. \name{} achieves 92.5\% accuracy in local-precision discrimination, substantially outperforming the best baseline at 70.0\%.
In the simulation environment, we evaluate \name{} (see Sec.~\ref{set:simu}) via online reinforcement learning. Specifically, 
our method achieves relative improvements of 10.9\% in InD success rates (0.674 v.s. 0.608) and 13.5\% in OOD success rates (0.353 v.s. 0.311), demonstrating its effectiveness in guiding policy learning.

%% file: document/related.tex
\section{Related Works}

\begin{figure*}[ht]
    \centering
        \includegraphics[width=\linewidth]{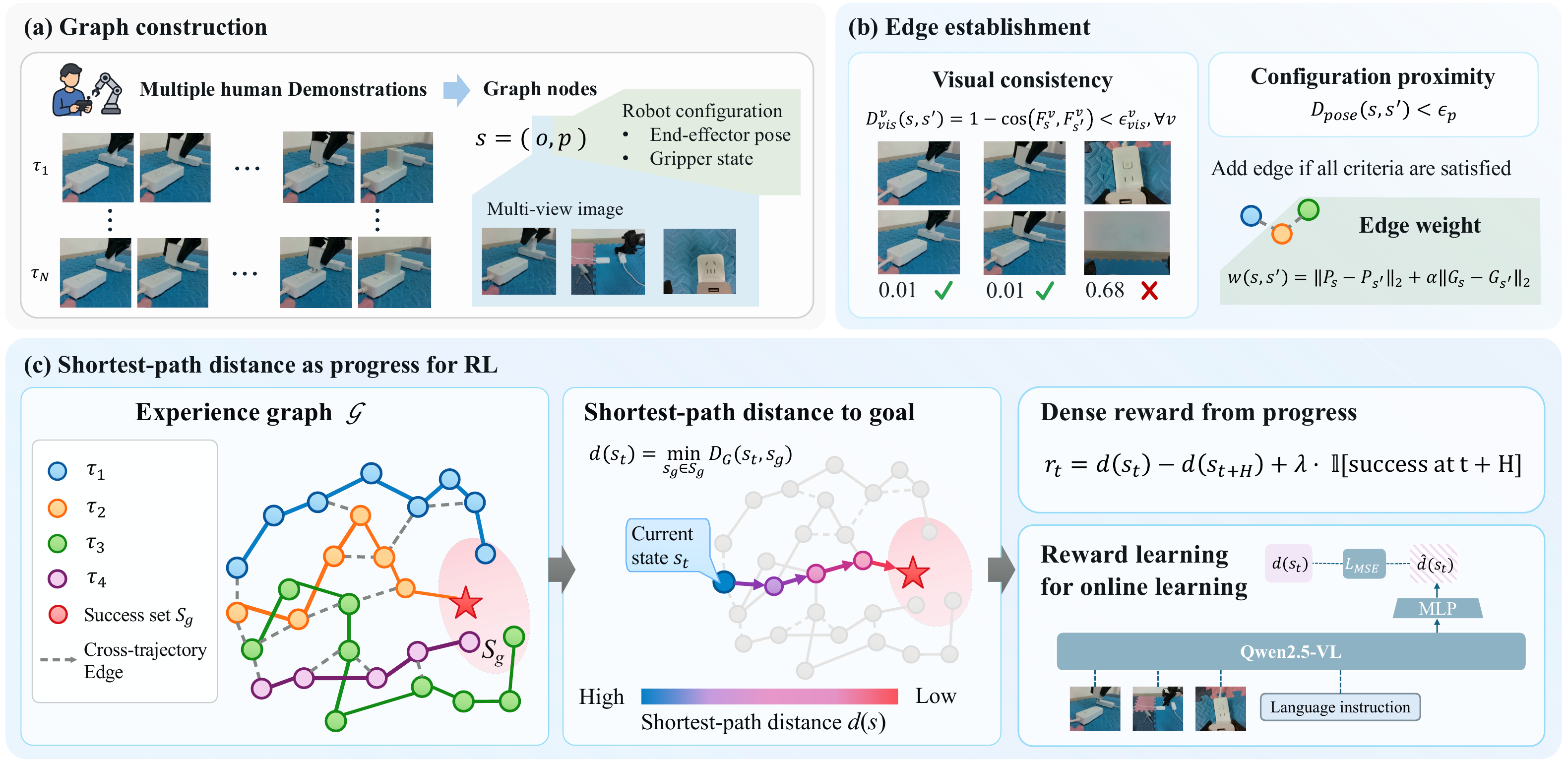} 

    \caption{{\textbf{An overview of \name{}.}} \textbf{(a): Graph construction.} Each demonstration timestep is represented by a graph node which comprises multi-view observations and the corresponding robot configuration.
    \textbf{(b): Edge establishment.} Cross-trajectory edges are established through multi-view visual consistency and configuration proximity, with weights reflecting pose and gripper-state differences. 
    \textbf{(c): Shortest-path distance as progress for RL.} The graph $\mathcal{G}$ defines the remaining distance to task completion as the weighted shortest-path distance from the current node to the goal. This distance serves as a measure of task progress.}
    \label{method}
\end{figure*}


\subsection{Progress estimation in robotics}
Obtaining an effective reward function is fundamental in robotic reinforcement learning to guide policy optimization and efficient exploration. This is especially crucial for high-precision tasks, where sparse feedback often leads to optimization collapse due to long credit-assignment horizons~\cite{xu2026rl}.
To derive dense rewards from demonstrations, existing methods commonly construct progress estimation from the temporal structure of trajectories.
Early approaches adopt linear temporal progress, such as normalized remaining-step estimation or linear interpolation over execution time~\cite{intelligence2025pi06vlalearnsexperience, lv2026vivavideogenerativevaluemodel, liang2026robometerscalinggeneralpurposerobotic, zhang2025rewind, zhai2025visionlanguageactioncriticmodelroboticrealworld}.
More recent approaches introduce structured, non-uniform progress estimation~\cite{tian2026position, mao2026armadvantagerewardmodeling}. SARM~\cite{chen2025sarmstageawarerewardmodeling} introduces subtask-level progress allocation with linear interpolation within each subtask, while ProcVLM~\cite{feng2026procvlmlearningproceduregroundedprogress} further refines it by redistributing intra-subtask progress according to visual changes. 
ARM~\cite{mao2026armadvantagerewardmodeling} learns from human-annotated progress directions, which incurs substantial time and labor costs.
Although these designs relax the uniform-rate assumption of linear interpolation, their progress targets are still largely derived from temporal ordering in demonstrations, enforcing progress to increase monotonically along the trajectory. Such temporal monotonicity does not necessarily reflect effective progress in high-precision manipulation. Demonstrations may contain hesitation, overshooting, or redundant adjustments where later observations can be temporally closer to task completion yet physically farther from a successful state.

\subsection{Learning precise manipulation}
Reinforcement learning in high-precision tasks has enabled remarkable breakthroughs, such as screw installation~\cite{xu2026rl}, shoelace tying~\cite{li2025gr}, in-hand manipulation~\cite{andrychowicz2020learning, chen2022system}, band tying~\cite{yin2024offline}, belt assembly~\cite{luo2025precise}, and deformable object handling~\cite{chi2024iterative}. 
While these works improve precise manipulation, reliable dense progress supervision from offline demonstrations remains unaddressed.
Beyond these methods, researchers are also exploring efficient strategies for learning high-precision tasks. 
Compared to these works that achieve high-precision manipulation with reweighted behavior cloning~\cite{yin2024offline, li2025gr} or pure reinforcement learning~\cite{li2025gr, andrychowicz2020learning, chen2022system, luo2025precise,xu2026rl}, PRIME distinguishes itself by focusing on accurate progress estimation learning, which further drives high-efficiency reinforcement learning.

%% file: document/method.tex
\section{Methodology}

\textbf{Overview.}\quad 
In this paper, we estimate task progress by aggregating collective experience across demonstrations through a weighted experience graph.
In Sec.~\ref{sec:challenge}, we first analyze the problem and formulate its solution as a graph structure. In Sec.~\ref{sec:neighbor} and Sec.~\ref{sec:dijkstra}, we further detail its construction and shortest-path computation. Sec.~\ref{sec:train} describes how the resulting progress estimates are used for policy learning. An overview is shown in Fig.~\ref{method}.

\subsection{Problem Analysis}
\label{sec:challenge}

Naturally, progress estimation can be formulated as a measure on representations of current world states, namely, $l_{\text{task}}(r(s_t), r(s_{t'}))$, where $l_{\text{task}}$ is a task-specific metric, and $r(s_t)$ is a representation of the current state.
In this formulation, the state representing function $r(\cdot)$ should be equipped with high distinguishing power, semantic awareness, and robustness to noise. To this end, we adopt DINOv3~\cite{simeoni2025dinov3}, a powerful self-supervised visual representation model that meets all the requirements~\cite{zhou2025dinowmworldmodelspretrained, edstedt2026romav2harderbetter}. It distinguishes visually similar yet semantically distant world states, such as peg rotation, dropping, insertion, etc. Note that $s_t$ can take a variety of forms, and here we formulate it as $s_t=(o_t,\mathbf{p}_t)$, where $o_t$ is the multi-view visual observation and $\mathbf{p}_t$ is the robot configuration. The robot configuration $\mathbf{p}_t$ is composed of the 3D positions of the TCP and the two gripper fingertips, which is denoted as $\mathbf{p}_t = [\mathbf{p}_t^{\mathrm{tcp}}, \mathbf{g}_t^{\mathrm{gripper}}]$. Although our work readily admits other forms of state representations such as $s_t=(o_{<t},\mathbf{p}_{<t},\mathbf{v}_{<t})$ as dictated by memory mechanisms~\cite{shi2026memoryvlaperceptualcognitivememoryvisionlanguageaction}, they are not the main focus of this paper and we leave them as future work.
Designing $l_{\text{task}}$ can be more subtle. While most existing works adopt a simple Euclidean distance in the representation space, e.g. $l_{\text{task}}(r(s_t), r(s_{t'})) = \|r(s_t)-r(s_{t'})\|_2$ and $r(s_t)=t$~\cite{intelligence2025pi06vlalearnsexperience,lv2026vivavideogenerativevaluemodel,liang2026robometerscalinggeneralpurposerobotic} or $r(s_t)=Enc(o_t)$~\cite{feng2026procvlmlearningproceduregroundedprogress}, we find the approach unsuitable for high-precision tasks as the latent paths indicated by Euclidean distances in the representation space of $r$ are often infeasible in the physical world. 
In such cases, geodesic distances along the feasible execution manifold are comparatively more appropriate. However, the feasible execution manifold is often unknown and difficult to model.
In this work, we propose to approximate the geodesic distance by Euclidean distance in neighborhood of representations.
Moreover, without a global prior regarding feasibility manifold, we propose to leverage the physical transitions collected in the dataset, and in this case, the minimum graph distance between states can be seen as a good approximation of the geodesic distance along the feasible execution manifold.

\subsection{Progress Estimator Construction}


\subsubsection{Nearest Neighbor State Retrieval}
\label{sec:neighbor}

To exploit collective experience across demonstrations, we construct graph connections from both observed transitions and state-compatible cross-trajectory transitions. 
Specifically, within each demonstration, we preserve the temporal structure of the trajectory by connecting consecutive states;
across demonstrations, we connect states only when they are close in robot configuration and consistent in visual task state, i.e., the following conditions must be satisfied:
\begin{equation}
\mathcal{N}(s) =
\left\{
s' \;\middle|\;
\lvert \mathbf{p}_{s} - \mathbf{p}_{s'} \rvert
\prec \boldsymbol{\epsilon}_{\mathrm{p}},
\quad D_{\mathrm{vis}}(s,s') < \epsilon_{\mathrm{vis}}
\right\}.
\label{eq:neighbor}
\end{equation}
We measure visual similarity by extracting DINOv3~\cite{simeoni2025dinov3} features $\mathbf F_s^v$ and \(\mathbf F_{s'}^v\) in each camera view \(v\), with \(D_{\mathrm{vis}}(s,s')=1-\cos(\mathbf F_s^v,\mathbf F_{s'}^v)\). Benefiting from the distinguishing power and strong pose-aware priors of DINOv3, physically irreversible transitions tend to result in higher latent distances, guaranteeing the physical transferability between cross-trajectory edges. 
We evaluate the similarity-based cross-trajectory edge forming mechanism in a real-world experiment. Specifically, we collect 20 new trajectories, compute the cross-trajectory edges between each new trajectory and the existing dataset (on average, more than 200 new edges per trajectory). For each trajectory, we replay the robotic arm to execute one-step transition between the new neighbor nodes formed by such cross-trajectory edges with environment unchanged. We find that over 99.4\% of these cross-trajectory edges are physically feasible, meaning that the connected states are mutually reachable through one-step transitions. False positives are observed only when the grasped object rotates by less than 3°.

\subsubsection{Shortest Path Planning via Dijkstra}
\label{sec:dijkstra}

We define the edge weight as $w(s_i,s_j)=
\left\|
\mathbf{p}_i^{\mathrm{tcp}}-\mathbf{p}_j^{\mathrm{tcp}}
\right\|_2
+
\alpha
\left\|
\mathbf{g}_i^{\mathrm{gripper}}-\mathbf{g}_j^{\mathrm{gripper}}
\right\|_2.
$ This retains fine configuration differences that matter for high-precision execution. After constructing a graph $\mathcal{G}=(V,E)$ from the visually gated neighborhood relation in Eq.~\eqref{eq:neighbor}, we apply Dijkstra's algorithm to compute the shortest path cost from state $s_t$ to the set of successful terminal states $s_g$, where \(s_g\) comprises terminal states identified by dataset success annotations or simulation-provided success signals:

\begin{equation}
d(s_t)=\min_{\pi\in\Pi(s_t,s_g)}
\sum_{(s_i,s_j)\in\pi}w(s_i,s_j).
\end{equation}
We denote the weighted distance along the shortest path from the current state $s_t$ to the goal state $s_g$ as $d(s_t)$. Benefiting from these designs, PRIME captures the variations in minimum feasible cost caused by suboptimal behaviors at a finer granularity, thereby benefiting real-world RL.

\subsection{Policy Learning}
\label{sec:train}

We use $d(s_t)$ as the metric for task progress. For an action chunk of length \(H\), we quantify its contribution to task progress by the reduction in remaining distance, \(r_t=d(s_t)-d(s_{t+H})\), where \(r_t>0\) indicates advancement toward task completion and \(r_t<0\) indicates a setback.

\paragraph{Learning Online} 
Upon constructing progress labels for the entire dataset, we train a distance estimator $\hat{d}_\phi(s_t,z)$ for online RL using MSE loss, where $z$ denotes the language instruction.
During online RL, the reward is defined as:

\begin{equation}
\begin{aligned}
\hat{r}^{\mathrm{on}}_t ={}&
\hat{d}_\phi(s_t,z)
-\hat{d}_\phi(s_{t+H},z) \\
&+\lambda_{\mathrm{succ}}
\mathbf{1}[\text{success at }t+H].
\end{aligned}
\end{equation}
Following \cite{zhang2025rewind}, we incorporate a success bonus into the final reward to encourage goal completion.

\paragraph{Learning Offline} 
Given a dataset annotated with progress estimation, we train the policy $\pi_\theta(a|o,z)$ using progress-prioritized sampling. Following the classic paradigm in RL~\cite{horgan2018distributed}, we construct a weighted sampler where the sampling probability $\text{Prob}(r_t)$ is determined by its filtered reward scaling $\text{Prob}(r_t) \propto \text{softmax} \left( \frac{\min(r_t, r_{\max})}{\tau} \right)$ . The training loss can be formulated as:
\begin{equation}
    L_\theta = -\mathbb{E}_{(s_t, z, a_{t:t+H}, r_t)\sim \text{Prob}(r_t) }\biggl[\log\pi_{\theta}(a_{t:t+H}|s_t,z)\biggl],
\end{equation}
where $r_{\max}$ clips rare high-reward outliers and $\tau>0$ controls the sampling concentration.


%% file: document/experiment.tex
\section{Experiments}

\input{document/main_result}

In this section, we conduct experiments to evaluate the accuracy of \name{}'s progress estimates and their effectiveness for offline policy learning and online reinforcement learning. We further examine the contributions of the distance metric and the sensitivity to graph and sampling parameters. Our aim is to address the following research questions:
\begin{itemize}
    \item[\textbf{Q1}] Can \name{} faithfully capture progress of high-precision tasks from human-collected trajectories interleaved with sub-optimal behaviors?
    \item[\textbf{Q2}] Can \name{}-based policy learn from sub-optimal trajectories in high-precision tasks?
    \item[\textbf{Q3}] Can \name{} provide more accurate supervision for progress estimation, thereby yielding more precise reward signals for RL training?
    \item[\textbf{Q4}] How do design choices and hyper-parameters affect \name{}?
\end{itemize}

\subsection{Experiment Setup}


\paragraph{Real Robot}

We evaluate PRIME on four real-world manipulation tasks using a Piper-X robot arm: drawer opening followed by object placement, peg-in-hole, peg-in-hole (Hard), and charger insertion (Fig.~\ref{fig:setting}). We collect 20, 30, 50, and 50 demonstration episodes for these tasks, respectively. These demonstrations are used to construct the progress estimator and subsequently train the policy. Each trained policy is evaluated over 30 randomized trials per task.

\paragraph{Simulation} 
\label{set:simu} 
Our simulation environment, termed ExpandedManiSkill, builds upon the standard configuration of RL4VLA~\cite{liu2026can}. Specifically, to verify the scalability of \name{} under substantial spatial variations and across a wide array of object geometries, we extend the initial positional workspace from $15\text{cm} \times 15\text{cm}$ to $15\text{cm} \times 40\text{cm}$ with more evaluation objects.


\paragraph{Baselines} We compare our approach against pure SFT and several advanced progress estimation baselines. In the real-world setting, we employ $\pi_{0.5}$~\cite{intelligence2025pi05visionlanguageactionmodelopenworld} as our base model. We further evaluate \name{} against RECAP~\cite{intelligence2025pi06vlalearnsexperience}, SARM~\cite{chen2025sarmstageawarerewardmodeling}, RoboMeter~\cite{liang2026robometerscalinggeneralpurposerobotic}, ViVa~\cite{lv2026vivavideogenerativevaluemodel}, GSR~\cite{yin2024offline} and ProcVLM~\cite{feng2026procvlmlearningproceduregroundedprogress}. In the simulated RL setting, we conduct comparative analyses against 0/1 sparse rewards and time-progress rewards. For fair comparison, the progress estimators mentioned above are trained using Qwen3-VL~\cite{bai2025qwen3vltechnicalreport} in real-world scenarios, and their estimated progress is subsequently used in the same progress-prioritized policy training pipeline. In simulation, Qwen2.5-VL~\cite{bai2025qwen25vltechnicalreport} is utilized as the progress-model backbone. During training, the model takes the robot configuration and observations as input.

\subsection{Can \name{} capture sub-optimal behaviors in human-collected dataset?}

\begin{figure*}[h]
    \centering
        \includegraphics[width=\linewidth]{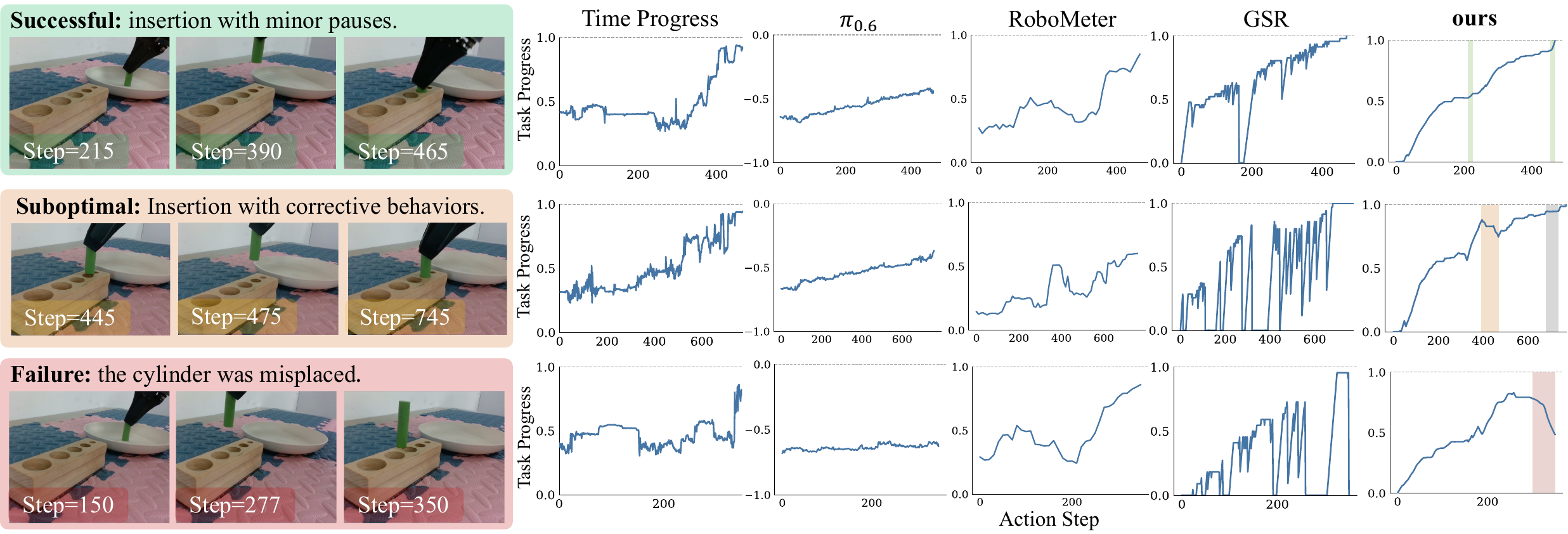} 
    \caption{\textbf{Evaluation on various robotic trajectories.} We compare our method against baselines across three scenarios. (1) Successful Execution: For a clean trajectory with minor pauses, \name{} provides a smooth, monotonic progress estimate. (2) Suboptimal Execution: When a human collector makes a mistake and lifts the cylinder upward to adjust, \name{} captures this setback with a local decrease in progress (orange band). (3) Failure Detection: When the cylinder is misplaced, \name{} reflects a sharp drop in progress (red band) as the trajectory drifts toward failure. For visualization, we normalize our distance predictions as \(1 - \frac{d(s_t)}{d(s_0)}\). More visualization results are provided in Appendix~\ref{app:vis}.}
    \label{fig:traj-eval}
\end{figure*}

We evaluate \name{} on Peg-in-hole (Hard) through qualitative progress visualization and a quantitative diagnostic benchmark. As illustrated in Fig.~\ref{fig:traj-eval}, the predicted curves show that \name{} captures setbacks during drifting and maintains stable estimates during idling. 
To further quantify estimation quality, a professional data collector collected an additional 60 trajectories exhibiting clearly suboptimal behavior and 40 expert trajectories. From these demonstrations, the collector selected 120 stage-comparison pairs with clear differences in task stage, along with 30 expert trajectories for assessing temporal smoothness, quantified by the normalized root mean square (RMS) of the second-order differences in predicted progress. The collector also carefully selected 120 local-precision comparison pairs exhibiting subtle yet task-critical geometric differences during insertion (e.g., observations taken just above the opening and those taken after suboptimal actions have caused horizontal drift).
We aim to evaluate whether a progress estimator can correctly capture task-stage transitions, reflect suboptimal behaviors through changes in estimated progress, and provide smooth progress signals for high-quality trajectories. These properties are important for producing informative and stable reward signals for downstream RL training.
As stated in Table~\ref{tab:diagnostic-benchmark}, \name{} achieves 92.5\% accuracy in local-precision discrimination, exceeding the best baseline by 22.5 percentage points while maintaining smooth progress predictions.

\begin{table}[t]
    \centering
    \caption{Results on the diagnostic benchmark.}
    \label{tab:diagnostic-benchmark}
    \begin{tabular}{lcccc}
        \toprule
        Method & \makecell{Stage-level\\Ordering $\uparrow$} & \makecell{Local\\Precision $\uparrow$} & \makecell{Progress\\Continuity $ (s^{-2}) \downarrow$} \\
        \midrule
        Linear & 73.3\% & 70.0\% & 8.14 \\
        SARM & 67.5\% & 67.5\% & 3.47 \\
        RoboMeter & 70.8\% & 61.7\% & 4.72 \\
        ViVa &  79.2\% & 40.8\% & \textbf{1.74} \\
        ProcVLM & 87.5\% & 67.5\% &  5.50\\
        RECAP & 98.3\% & 68.3\% & 2.69 \\
        GSR & 79.2\% & 65.8\% & 16.01 \\
        \textbf{\name{}} & \textbf{100.0\%} & \textbf{92.5\%} & 1.78 \\
        \bottomrule
    \end{tabular}
\end{table}

\subsection{Can PRIME-based policy learn from sub-optimal trajectories in high-precision tasks?}

We consider an offline setting with two types of datasets: noisy data containing suboptimal trajectories (e.g., multiple insertion attempts), and expert data comprising high-quality demonstrations with inevitable human errors for four tasks: drawer opening followed by object placement, peg-in-hole, peg-in-hole (Hard), and charger insertion. The results in Table~\ref{tab:main} illustrate that \name{} consistently and remarkably outperforms the advanced progress estimation methods.

\subsection{Can \name{} enhance progress modeling and thereby facilitate online RL training?}

In this subsection, we evaluate \name{} on the ExpandedManiSkill (See Sec~\ref{set:simu}). To mimic the inherent errors made by human operators in real-world data collection, we introduce perturbations into the dataset~\cite{zhang2025rewind} to replicate suboptimal behaviors in real world. This perturbed dataset is subsequently utilized to train the progress estimator. Following prior works~\cite{zhang2025rewind}, we trained the RL agent for 2M steps and report the average evaluation results over the final 1M steps, with evaluations every 25k steps.

\begin{table}[h]
    \centering
    \caption{\textbf{Main results from the simulation.}  We report the average success rates across 3 seeds (0, 1 and 2).}
    \begin{tabular}{c|cc}
        \toprule
        \multirow{2}{*}{\makecell[c]{Reward\\Model Type}} & \multicolumn{2}{c}{Success Rate} \\
        & In-distribution & Out-of-distribution \\ 
        \midrule
        Sparse reward & $54.0{\scriptstyle \pm 2.3}$ & $27.1{\scriptstyle \pm 1.4}$ \\
        Time Progress & $60.8{\scriptstyle \pm 6.1}$ & $31.1{\scriptstyle \pm 1.5}$ \\
        Ours & $\mathbf{67.4}{\scriptstyle \pm 5.6}$ & $\mathbf{35.3}{\scriptstyle \pm 3.1}$ \\
        \bottomrule
    \end{tabular}
\end{table}

\subsection{Effects of Design Choices and Hyper-parameters}

\label{sec:abla}

In this subsection, we conduct ablation to investigate the critical design elements in \name{} on Peg-Hard and Drawer.

\paragraph{Ablation on Graph Construction and Progress Distance}

We ablate the key components of \name{} to examine the roles of graph connectivity and distance computation. 
As illustrated in Table~\ref{tab:abla-distance}, \name{} substantially outperforms all three variants.  
\name{} w/o geometric edge costs retains the same graph connectivity but assigns unit edge costs, ignoring geometric transition magnitudes, thereby measuring only the minimum number of steps to the goal. \name{} w/o configuration-proximity constraint removes the configuration-proximity constraint when establishing cross-trajectory edges, relying solely on visual similarity while keeping the remaining graph formulation unchanged. This may connect visually similar states despite substantial differences in robot configuration, potentially introducing inappropriate cross-trajectory shortcuts. The Euclidean distance to goal discards the graph structure, potentially conflating equally distant states even at different task stages. 
These results demonstrate the complementary importance of configuration-aware connectivity, fine-grained transition costs, and graph-constrained path structure. 

\begin{table}[h]
\centering
\caption{Ablation of graph connectivity and distance computation in \name{}.}
\label{tab:abla-distance}

\begin{tabular}{lcc}
\toprule
Method & Peg-Hard & Drawer \\
\midrule
\name{} w/o geometric edge costs            & 33.3\% & 90.0\% \\
\name{} w/o configuration-proximity constraint      & 26.7\% & 53.3\% \\
Euclidean distance to goal (no graph)& 56.7\% & 33.3\% \\
\name{}                              & \textbf{70.0\%} & \textbf{96.7\%} \\
\bottomrule
\end{tabular}
\end{table}

\begin{figure}[h]
    \centering
    \includegraphics[width=0.49\linewidth]{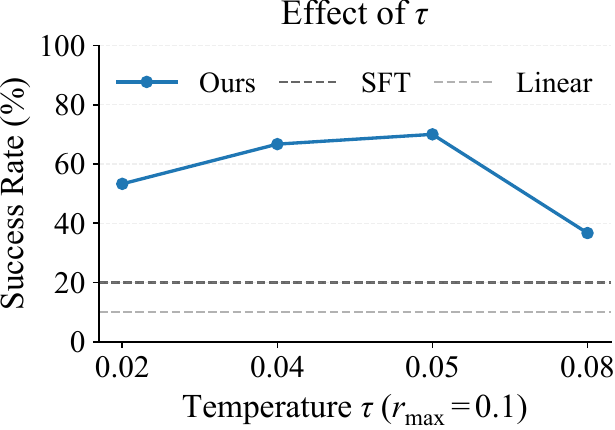}\hfill
    \includegraphics[width=0.49\linewidth]{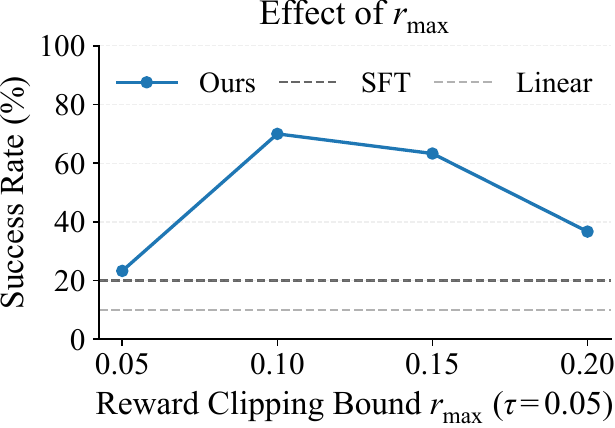}
    \caption{Effect of temperature $\tau$ (left) and reward clipping bound $r_{\max}$ (right).}
    \label{fig:ablation-tau-rmax}
\end{figure}

\paragraph{Effect of Hyper-parameter Reward Clipping Bound $r_{\max}$ and Temperature $\tau$} To identify the joint effects of the core hyper-parameters governing our filtered sampling distribution, we evaluate the policy performance by varying the reward clipping bound $r_{\max}$ and softmax temperature $\tau$ in Fig.~\ref{fig:ablation-tau-rmax}. The results show that our method consistently outperforms both SFT and Linear across all evaluated settings.

\paragraph{Effect of Hyper-parameter Neighbor Distance Threshold $\epsilon$} 
The selection of the threshold depends on the differences in robot configurations between temporally adjacent frames. For instance, $P_{70}$ implies that the filtering of the neighbor set is based on the 70th percentile of the configuration differences between these temporal neighbors.
As illustrated in Table~\ref{tab:percentile_sr}, within a certain range of thresholds, \name{} consistently outperforms the baseline.
However, being too conservative restricts cross-trajectory connections and drops the success rate to $30\%$, while being overly aggressive introduces invalid bridges, with a success rate of $53.3\%$. 

\begin{table}[h]
\centering
\caption{Success rate under different percentile choices for neighbor distance thresholds $\epsilon_{xy}$ and $\epsilon_{z}$}
\label{tab:percentile_sr}
\begin{tabular}{c|ccc}
\toprule
\diagbox{$\epsilon_{xy}$}{$\epsilon_{z}$} & $P_{30}$ & $P_{50}$ & $P_{70}$ \\ \midrule
$P_{50}$  & $30.0\%$        & $43.3\%$        &   $66.7\%$      \\ 
$P_{90}$  & $36.7\%$   & \textbf{$70.0\%$}   & $66.7\%$   \\ 
$P_{100}$ & $53.3\%$   & $60.0\%$   & $53.3\%$   \\ 
\bottomrule
\end{tabular}
\end{table}

\begin{figure}[h]
    \centering
    \begin{minipage}[c]{0.44\linewidth}
        \centering
        \includegraphics[width=\linewidth]{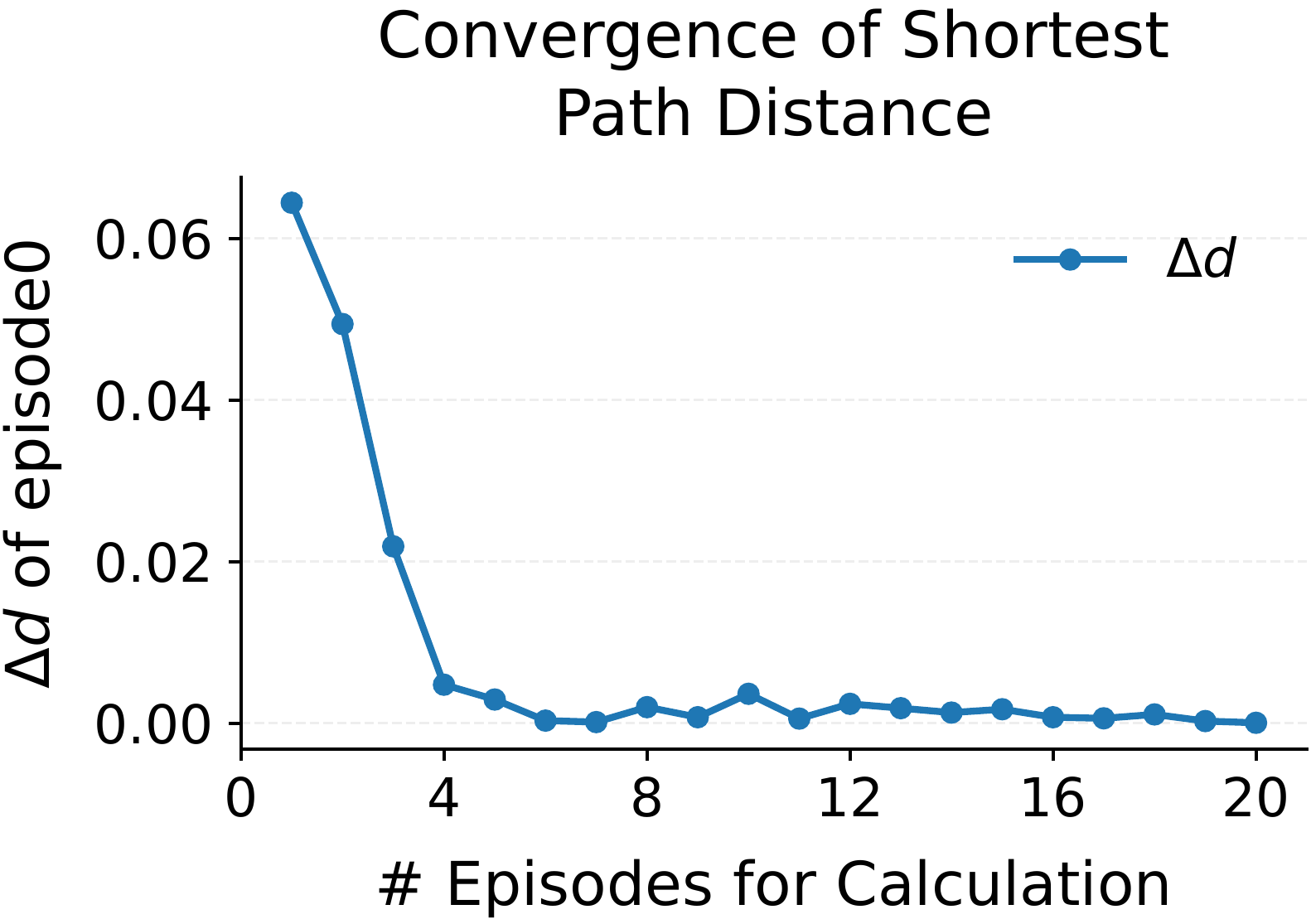}
    \end{minipage}\hfill
    \begin{minipage}[c]{0.54\linewidth}
        \centering
        \fontsize{8pt}{9pt}\selectfont
        \setlength{\tabcolsep}{2pt}
        \begin{tabular}{lcc}
            \toprule
            \#Frames & \makecell{Visual\\Encode (s)} & \makecell{Graph\\Construction (s)} \\
            \midrule
            25,000  & 249.1   & 0.40   \\
            50,000  & 492.4   & 1.13   \\
            75,000  & 741.6   & 1.91   \\
            150,000 & 1,521.5 & 5.88  \\
            300,000 & 2,893.0 & 19.08 \\
            \bottomrule
        \end{tabular}
    \end{minipage}

    \caption{\textbf{Convergence and computational efficiency.}
    (Left) The shortest-path deviation $\Delta d$ stabilizes after 4 episodes.
    (Right) Computation time of visual encoding and graph construction with increasing numbers of graph nodes. All runtime measurements are obtained on an NVIDIA RTX3090 GPU.}
    \label{fig:convergence}
\end{figure}

\paragraph{Convergence Efficiency and Computational Efficiency} We evaluate the convergence and computational efficiency of our graph construction. As shown in Fig.~\ref{fig:convergence}, the shortest-path deviation $\Delta d$ for Episode 0 rapidly decreases and stabilizes near zero after only four episodes. 
Fig.~\ref{fig:convergence} (right) further shows that, with 300,000 frames, graph construction takes 19.08s, while visual encoding takes 2,893.0s. This indicates that incorporating the graph representation adds only a small computational cost while maintaining efficient scalability to long sequences.

%% file: document/main_result.tex
\begin{table*}[ht]
\caption{\textbf{Main results from the real experiments.} We evaluate the performance of \name{} across four robotic manipulation tasks of varying complexity: Drawer Opening, Peg-in-Hole (Standard and Hard), and Charger Insertion. These evaluations are conducted under diverse data quality conditions, encompassing noisy data that is unqualified for pure imitation learning, as well as expert data that is qualified for training but inevitably contains human mistakes. Experimental results demonstrate that \name{} consistently outperforms standard baselines across various levels of data quality. Besides, the gains on expert data suggest that local progress information remains useful even when overall demonstration quality is high.}
\label{tab:main}
\centering
\fontsize{9pt}{10pt}\selectfont
\setlength{\tabcolsep}{4pt}
\begin{tabular}{ll|c|cccc|cccc}
\toprule
\multirow{4}{*}{Task} & \multirow{4}{*}{Demo} & \multicolumn{8}{c}{Success Rate} \\
\cmidrule(lr){3-11}
& & \multicolumn{1}{c|}{\makecell{Behavior\\Cloning}}
& \multicolumn{4}{c|}{Linear Temporal Progress}
& \multicolumn{4}{c}{\makecell{Non-uniform Progress}}\\
\cmidrule(lr){3-3}\cmidrule(lr){4-7}\cmidrule(lr){8-11}
& & $\pi_{0.5}$-SFT & Linear & RECAP & ViVa & RoboMeter & SARM & GSR-based & ProcVLM & \textbf{Ours} \\
\midrule
\multirow{2}{*}{Drawer} & Noisy & 86.7\% & 80.0\% & 90.0\% & 90.0\% & 93.3\% & 90.0\% & 46.7\% & 90.0\% & \textbf{96.7\%} \\
& Expert & \textbf{100.0\%} & 90.0\% & \textbf{100.0\%} & \textbf{100.0\%} & \textbf{100.0\%} & \textbf{100.0\%} & 86.7\% & \textbf{100.0\%} & \textbf{100.0\%} \\
\midrule
\multirow{2}{*}{Peg} & Noisy & 33.3\% & 16.7\% & 40.0\% & 43.3\% & 50.0\% & 53.3\% & 26.7\% & 46.7\% & \textbf{73.3\%} \\
& Expert & 86.7\% & 76.7\% & 93.3\% & 86.7\% & 90.0\% & 73.3\% & 73.3\% & 86.7\% & \textbf{100.0\%} \\
\midrule
\multirow{2}{*}{Peg-Hard} & Noisy & 20.0\% & 10.0\% & 30.0\% & 36.7\% & 36.7\% & 23.3\% & 13.3\% & 43.3\% & \textbf{70.0\%} \\
& Expert & 66.7\% & 33.3\% & 56.7\% & 53.3\% & 90.0\% & 46.7\% & 40.0\% & 66.7\% & \textbf{100.0\%} \\
\midrule
\multirow{2}{*}{Charger} & Noisy & 10.0\% & 0.0\% & 30.0\% & 33.3\% & 30.0\% & 26.7\% & 20.0\% & 33.3\% & \textbf{50.0\%} \\
& Expert & 46.7\% & 26.7\% & 36.7\% & 33.3\% & 23.3\% & 23.3\% & 16.7\% & 26.7\% & \textbf{66.7\%} \\
\bottomrule
\end{tabular}
\end{table*}

%% file: document/conclusion.tex
\vspace{-2mm}
\section{Conclusion}

In this paper, we introduce an improved paradigm for task progress estimation and policy learning in robotic manipulation, particularly for high-precision tasks. By combining human-collected trajectories into a graph-based framework, \name{} provides an estimate of the geodesic distance and treats it as a metric for progress estimation and as a reward for reinforcement learning. Based on such reward-labeling paradigm, our extensive evaluations on ExpandedManiSkill and in real-world deployments, comprising peg-in-hole, high-precision peg-in-hole, and charger insertion, demonstrate significant improvements in success rates (up to 5$\times$ for charger insertion and 3.5$\times$ for high-precision peg-in-hole). Furthermore, \name{} introduces modest computational overhead for graph construction and improves performance under the evaluated out-of-distribution conditions.

%% file: document/appendix.tex
\label{appendix}

\subsection{Real-World Setup}

\begin{figure}[h]
    \centering
        \includegraphics[width=\linewidth]{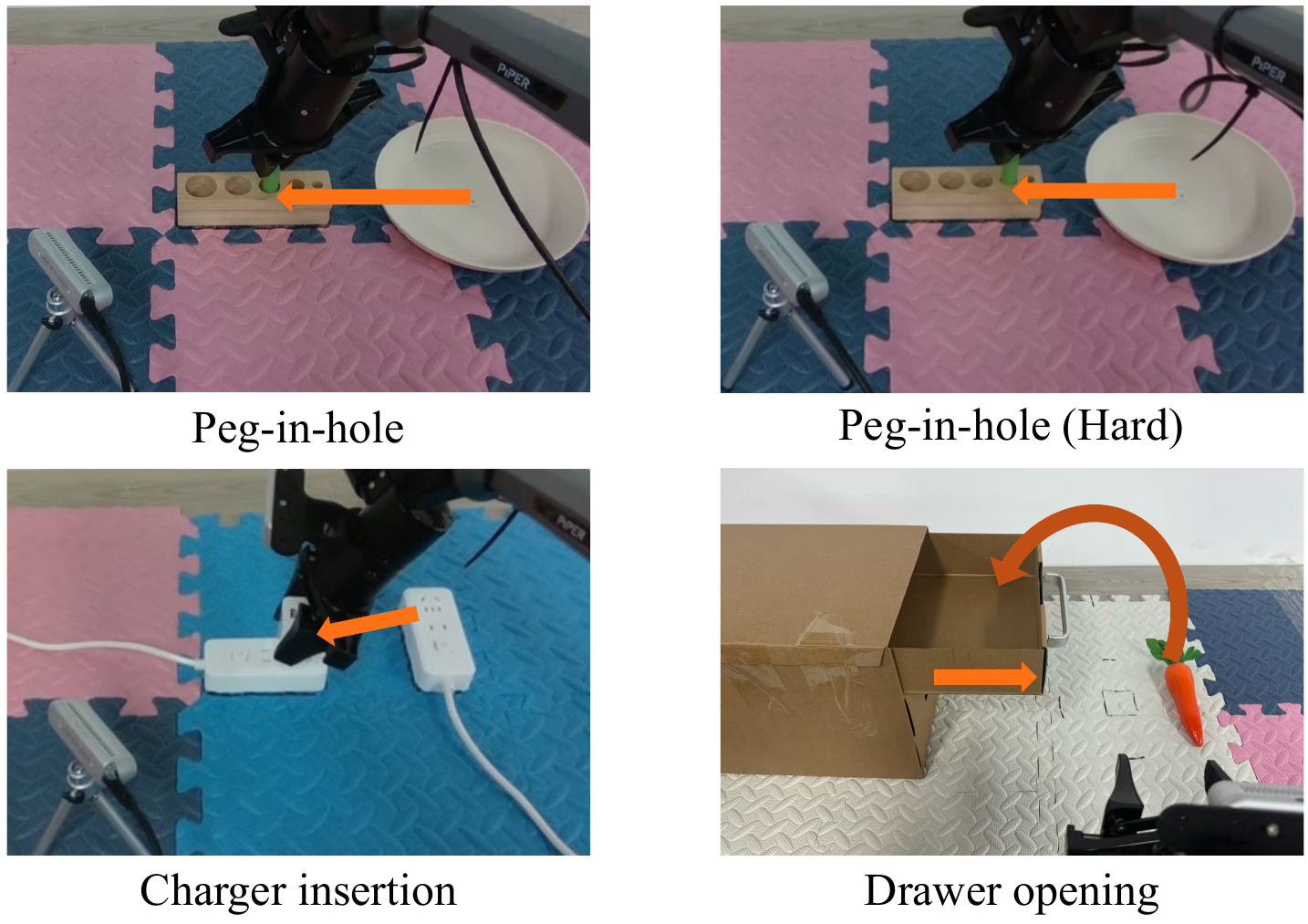}
    \caption{An overview of our real-world experimental setup.}
    \label{fig:setting}

\end{figure}

\subsection{Implementation Details}

We provide a comprehensive overview of the hyperparameter configurations in our experiments.

\begin{table}[h]
\caption{Training Details of \name{}.}
\centering
\begin{tabular}{c|c|c|c}
\toprule
\multicolumn{4}{c}{Hyper-parameters} \\
\midrule
\multicolumn{2}{c|}{Action Prediction Horizon}
& \multicolumn{2}{c}{50} \\
\multicolumn{2}{c|}{Action Horizon}
& \multicolumn{2}{c}{50} \\
\multicolumn{2}{c|}{Observation Horizon}
& \multicolumn{2}{c}{1} \\
\multicolumn{2}{c|}{Gradient Steps}
& \multicolumn{2}{c}{80000} \\
\multicolumn{2}{c|}{Batch Size}
& \multicolumn{2}{c}{16} \\
\midrule
$r_{\max}$ &0.1 & $\epsilon_{vis}$ & 0.1 \\
$\tau$ & 0.05 & $\alpha$ & 0.1 \\
$\epsilon_{xy}$ & $P_{90}$ & $\lambda_{succ}$ & 1 \\
$\epsilon_z$ & $P_{50}$ & learning rate & $1e-5$\\
\bottomrule
\end{tabular}
\vspace{-3mm}
\end{table}



\subsection{Progress Estimation Visualization}

\label{app:vis}

In this subsection, we present more visualization examples of our progress estimation predictions.



\begin{figure}[h]
    \centering
        \includegraphics[width=\linewidth]{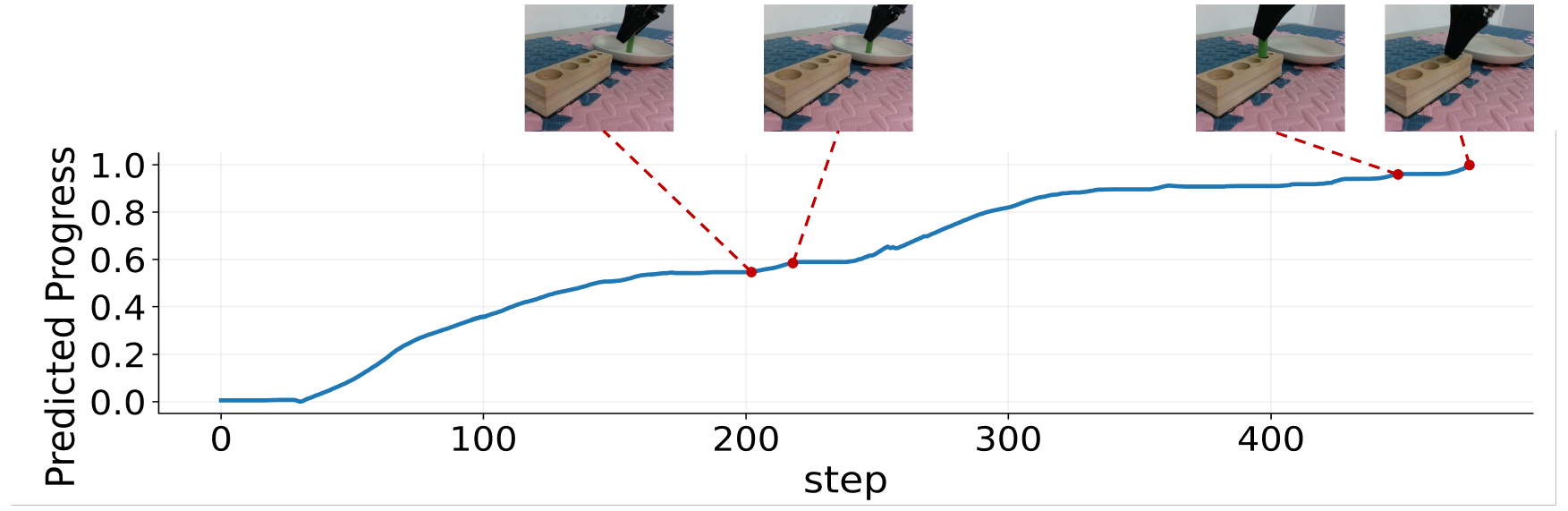}
    \caption{Expert demonstrations with pauses in peg-in-hole (Hard) task.}
    \label{fig:app1}

\end{figure}

\begin{figure}[h]
    \centering
        \includegraphics[width=\linewidth]{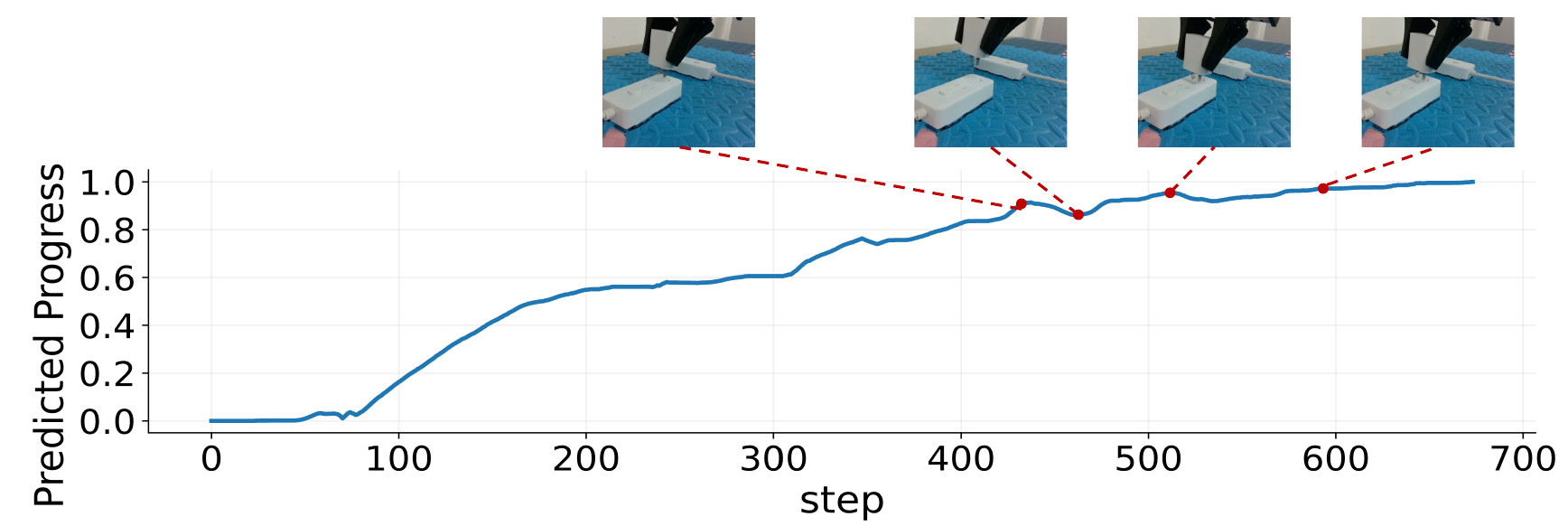}
    \caption{Suboptimal behaviors in charger insertion task.}
    \label{fig:app1}

\end{figure}

\begin{figure}[h]
    \centering
        \includegraphics[width=\linewidth]{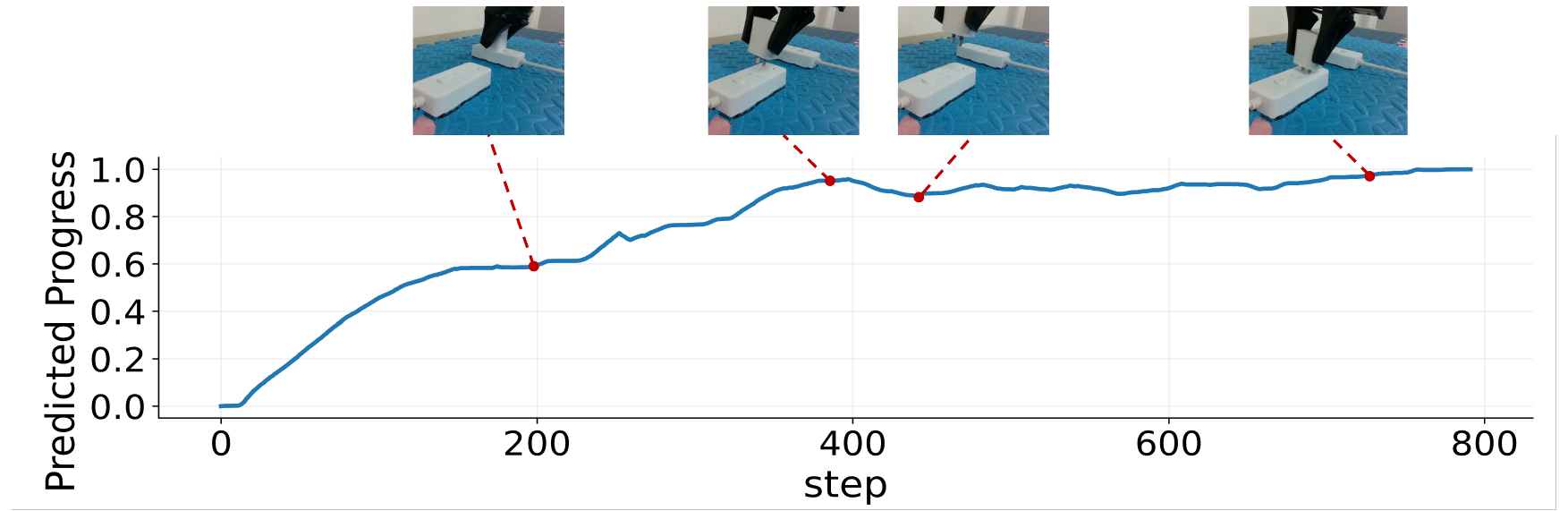}
    \caption{Suboptimal behaviors in charger insertion task.}
    \label{fig:app1}

\end{figure}

\begin{figure}[h]
    \centering
        \includegraphics[width=\linewidth]{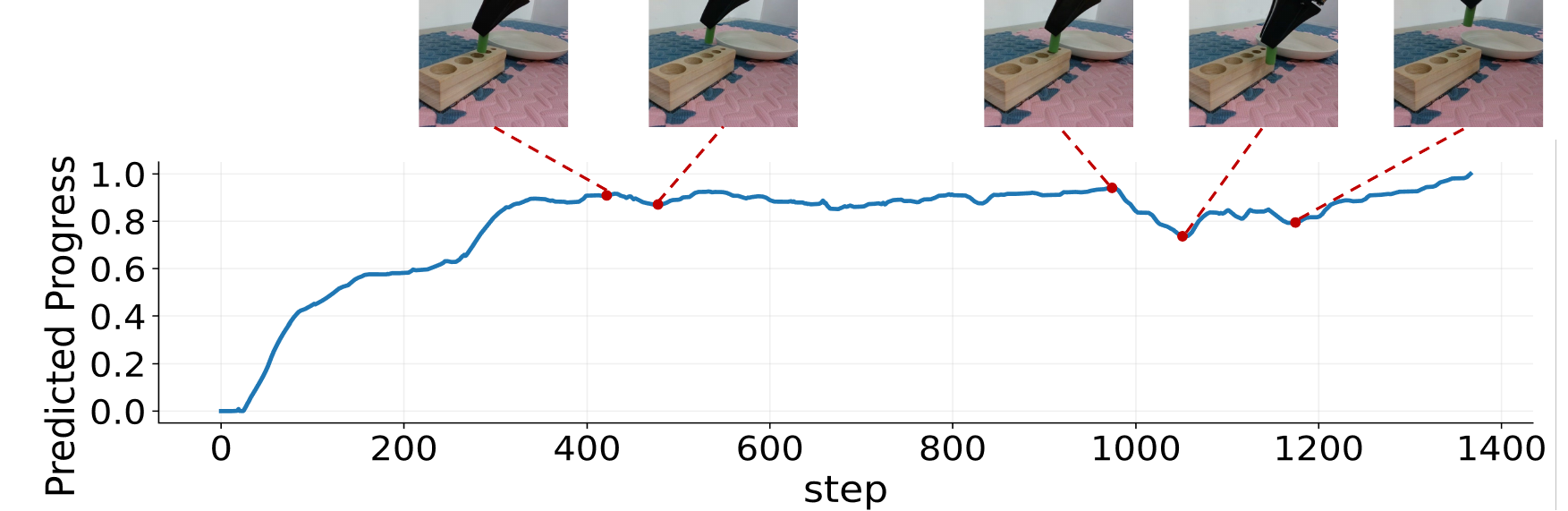}
    \caption{Demonstrations from an untrained human, exhibiting severe boundary violations.}
    \label{fig:app1}

\end{figure}



\begin{figure}[h]
    \centering
        \includegraphics[width=\linewidth]{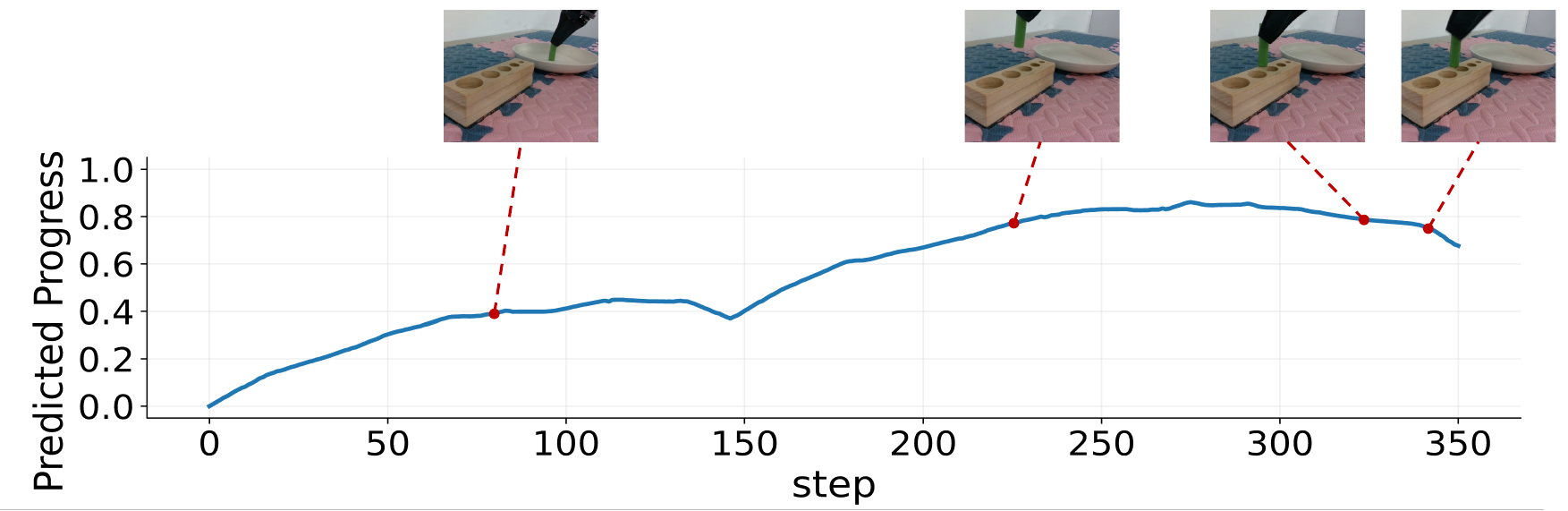}
    \caption{The cylinder is placed in an incorrect location.}
    \label{fig:app1}

\end{figure}

%% file: custom.bib
@article{liu2026can,
  title={What can rl bring to vla generalization? an empirical study},
  author={Liu, Jijia and Gao, Feng and Wei, Bingwen and Chen, Xinlei and Liao, Qingmin and Wu, Yi and Yu, Chao and Wang, Yu},
  journal={Advances in Neural Information Processing Systems},
  volume={38},
  pages={97121--97151},
  year={2026}
}

@misc{intelligence2025pi06vlalearnsexperience,
      title={$\pi^{*}_{0.6}$: a VLA That Learns From Experience}, 
      author={Physical Intelligence and Ali Amin and Raichelle Aniceto and Ashwin Balakrishna and Kevin Black and Ken Conley and Grace Connors and James Darpinian and Karan Dhabalia and Jared DiCarlo and Danny Driess and Michael Equi and Adnan Esmail and Yunhao Fang and Chelsea Finn and Catherine Glossop and Thomas Godden and Ivan Goryachev and Lachy Groom and Hunter Hancock and Karol Hausman and Gashon Hussein and Brian Ichter and Szymon Jakubczak and Rowan Jen and Tim Jones and Ben Katz and Liyiming Ke and Chandra Kuchi and Marinda Lamb and Devin LeBlanc and Sergey Levine and Adrian Li-Bell and Yao Lu and Vishnu Mano and Mohith Mothukuri and Suraj Nair and Karl Pertsch and Allen Z. Ren and Charvi Sharma and Lucy Xiaoyang Shi and Laura Smith and Jost Tobias Springenberg and Kyle Stachowicz and Will Stoeckle and Alex Swerdlow and James Tanner and Marcel Torne and Quan Vuong and Anna Walling and Haohuan Wang and Blake Williams and Sukwon Yoo and Lili Yu and Ury Zhilinsky and Zhiyuan Zhou},
      year={2025},
      eprint={2511.14759},
      archivePrefix={arXiv},
      primaryClass={cs.LG},
      url={https://arxiv.org/abs/2511.14759}, 
}

@inproceedings{zhang2025rewind,
  title={ReWiND: Language-Guided Rewards Teach Robot Policies without New Demonstrations},
  author={Zhang, Jiahui and Luo, Yusen and Anwar, Abrar and Sontakke, Sumedh Anand and Lim, Joseph J and Thomason, Jesse and Biyik, Erdem and Zhang, Jesse},
  booktitle={Conference on Robot Learning},
  pages={460--488},
  year={2025},
  organization={PMLR}
}

@techreport{tian2026position,
  title   = {Position: Good Embodied Reward Models Need Bad Behavior Data},
  author  = {Ran Tian and Yilin Wu and Andrea Bacjsy},
  year    = {2026},
  institution = {Carnegie Mellon University},
  url     = {https://cmu-intentlab.github.io/pdf/tian_icml_26_position.pdf}
}

@misc{liang2026robometerscalinggeneralpurposerobotic,
      title={Robometer: Scaling General-Purpose Robotic Reward Models via Trajectory Comparisons}, 
      author={Anthony Liang and Yigit Korkmaz and Jiahui Zhang and Minyoung Hwang and Abrar Anwar and Sidhant Kaushik and Aditya Shah and Alex S. Huang and Luke Zettlemoyer and Dieter Fox and Yu Xiang and Anqi Li and Andreea Bobu and Abhishek Gupta and Stephen Tu and Erdem Biyik and Jesse Zhang},
      year={2026},
      eprint={2603.02115},
      archivePrefix={arXiv},
      primaryClass={cs.RO},
      url={https://arxiv.org/abs/2603.02115}, 
}

@misc{chen2025sarmstageawarerewardmodeling,
      title={SARM: Stage-Aware Reward Modeling for Long Horizon Robot Manipulation}, 
      author={Qianzhong Chen and Justin Yu and Mac Schwager and Pieter Abbeel and Yide Shentu and Philipp Wu},
      year={2025},
      eprint={2509.25358},
      archivePrefix={arXiv},
      primaryClass={cs.RO},
      url={https://arxiv.org/abs/2509.25358}, 
}

@misc{zhai2025visionlanguageactioncriticmodelroboticrealworld,
      title={A Vision-Language-Action-Critic Model for Robotic Real-World Reinforcement Learning}, 
      author={Shaopeng Zhai and Qi Zhang and Tianyi Zhang and Fuxian Huang and Haoran Zhang and Ming Zhou and Shengzhe Zhang and Litao Liu and Sixu Lin and Jiangmiao Pang},
      year={2025},
      eprint={2509.15937},
      archivePrefix={arXiv},
      primaryClass={cs.RO},
      url={https://arxiv.org/abs/2509.15937}, 
}

@misc{intelligence2025pi05visionlanguageactionmodelopenworld,
      title={$\pi_{0.5}$: a Vision-Language-Action Model with Open-World Generalization}, 
      author={Physical Intelligence and Kevin Black and Noah Brown and James Darpinian and Karan Dhabalia and Danny Driess and Adnan Esmail and Michael Equi and Chelsea Finn and Niccolo Fusai and Manuel Y. Galliker and Dibya Ghosh and Lachy Groom and Karol Hausman and Brian Ichter and Szymon Jakubczak and Tim Jones and Liyiming Ke and Devin LeBlanc and Sergey Levine and Adrian Li-Bell and Mohith Mothukuri and Suraj Nair and Karl Pertsch and Allen Z. Ren and Lucy Xiaoyang Shi and Laura Smith and Jost Tobias Springenberg and Kyle Stachowicz and James Tanner and Quan Vuong and Homer Walke and Anna Walling and Haohuan Wang and Lili Yu and Ury Zhilinsky},
      year={2025},
      eprint={2504.16054},
      archivePrefix={arXiv},
      primaryClass={cs.LG},
      url={https://arxiv.org/abs/2504.16054}, 
}

@misc{lv2026vivavideogenerativevaluemodel,
      title={ViVa: A Video-Generative Value Model for Robot Reinforcement Learning}, 
      author={Jindi Lv and Hao Li and Jie Li and Yifei Nie and Fankun Kong and Yang Wang and Xiaofeng Wang and Zheng Zhu and Chaojun Ni and Qiuping Deng and Hengtao Li and Jiancheng Lv and Guan Huang},
      year={2026},
      eprint={2604.08168},
      archivePrefix={arXiv},
      primaryClass={cs.RO},
      url={https://arxiv.org/abs/2604.08168}, 
}

@article{xu2026rl,
  title={RL Token: Bootstrapping Online RL with Vision-Language-Action Models},
  author={Xu, Charles and Springenberg, Jost Tobias and Equi, Michael and Amin, Ali and Esmail, Adnan and Levine, Sergey and Ke, Liyiming},
  journal={arXiv preprint arXiv:2604.23073},
  year={2026}
}

@inproceedings{horgan2018distributed,
  title={Distributed Prioritized Experience Replay},
  author={Horgan, Dan and Quan, John and Budden, David and Barth-Maron, Gabriel and Hessel, Matteo and van Hasselt, Hado and Silver, David},
  booktitle={International Conference on Learning Representations},
  year={2018}
}

@inproceedings{mavip,
  title={VIP: Towards Universal Visual Reward and Representation via Value-Implicit Pre-Training},
  author={Ma, Yecheng Jason and Sodhani, Shagun and Jayaraman, Dinesh and Bastani, Osbert and Kumar, Vikash and Zhang, Amy},
  booktitle={The Eleventh International Conference on Learning Representations}
}

@misc{bai2025qwen25vltechnicalreport,
      title={Qwen2.5-VL Technical Report}, 
      author={Shuai Bai and Keqin Chen and Xuejing Liu and Jialin Wang and Wenbin Ge and Sibo Song and Kai Dang and Peng Wang and Shijie Wang and Jun Tang and Humen Zhong and Yuanzhi Zhu and Mingkun Yang and Zhaohai Li and Jianqiang Wan and Pengfei Wang and Wei Ding and Zheren Fu and Yiheng Xu and Jiabo Ye and Xi Zhang and Tianbao Xie and Zesen Cheng and Hang Zhang and Zhibo Yang and Haiyang Xu and Junyang Lin},
      year={2025},
      eprint={2502.13923},
      archivePrefix={arXiv},
      primaryClass={cs.CV},
      url={https://arxiv.org/abs/2502.13923}, 
}

@misc{bai2025qwen3vltechnicalreport,
      title={Qwen3-VL Technical Report}, 
      author={Shuai Bai and Yuxuan Cai and Ruizhe Chen and Keqin Chen and Xionghui Chen and Zesen Cheng and Lianghao Deng and Wei Ding and Chang Gao and Chunjiang Ge and Wenbin Ge and Zhifang Guo and Qidong Huang and Jie Huang and Fei Huang and Binyuan Hui and Shutong Jiang and Zhaohai Li and Mingsheng Li and Mei Li and Kaixin Li and Zicheng Lin and Junyang Lin and Xuejing Liu and Jiawei Liu and Chenglong Liu and Yang Liu and Dayiheng Liu and Shixuan Liu and Dunjie Lu and Ruilin Luo and Chenxu Lv and Rui Men and Lingchen Meng and Xuancheng Ren and Xingzhang Ren and Sibo Song and Yuchong Sun and Jun Tang and Jianhong Tu and Jianqiang Wan and Peng Wang and Pengfei Wang and Qiuyue Wang and Yuxuan Wang and Tianbao Xie and Yiheng Xu and Haiyang Xu and Jin Xu and Zhibo Yang and Mingkun Yang and Jianxin Yang and An Yang and Bowen Yu and Fei Zhang and Hang Zhang and Xi Zhang and Bo Zheng and Humen Zhong and Jingren Zhou and Fan Zhou and Jing Zhou and Yuanzhi Zhu and Ke Zhu},
      year={2025},
      eprint={2511.21631},
      archivePrefix={arXiv},
      primaryClass={cs.CV},
      url={https://arxiv.org/abs/2511.21631}, 
}

@article{yin2024offline,
  title={Offline imitation learning through graph search and retrieval},
  author={Yin, Zhao-Heng and Abbeel, Pieter},
  journal={arXiv preprint arXiv:2407.15403},
  year={2024}
}

@article{andrychowicz2020learning,
  title={Learning dexterous in-hand manipulation},
  author={Andrychowicz, OpenAI: Marcin and Baker, Bowen and Chociej, Maciek and Jozefowicz, Rafal and McGrew, Bob and Pachocki, Jakub and Petron, Arthur and Plappert, Matthias and Powell, Glenn and Ray, Alex and others},
  journal={The International Journal of Robotics Research},
  volume={39},
  number={1},
  pages={3--20},
  year={2020},
  publisher={SAGE Publications Sage UK: London, England}
}

@inproceedings{chen2022system,
  title={A system for general in-hand object re-orientation},
  author={Chen, Tao and Xu, Jie and Agrawal, Pulkit},
  booktitle={Conference on robot learning},
  pages={297--307},
  year={2022},
  organization={PMLR}
}

@article{chi2024iterative,
  title={Iterative residual policy: for goal-conditioned dynamic manipulation of deformable objects},
  author={Chi, Cheng and Burchfiel, Benjamin and Cousineau, Eric and Feng, Siyuan and Song, Shuran},
  journal={The International Journal of Robotics Research},
  volume={43},
  number={4},
  pages={389--404},
  year={2024},
  publisher={SAGE Publications Sage UK: London, England}
}

@article{li2025gr,
  title={Gr-rl: Going dexterous and precise for long-horizon robotic manipulation},
  author={Li, Yunfei and Ma, Xiao and Xu, Jiafeng and Cui, Yu and Cui, Zhongren and Han, Zhigang and Huang, Liqun and Kong, Tao and Liu, Yuxiao and Niu, Hao and others},
  journal={arXiv preprint arXiv:2512.01801},
  year={2025}
}

@article{chen2025rm,
  title={RM-RL: Role-Model Reinforcement Learning for Precise Robot Manipulation},
  author={Chen, Xiangyu and Zhou, Chuhao and Liu, Yuxi and Yang, Jianfei},
  journal={arXiv preprint arXiv:2510.15189},
  year={2025}
}

@article{huang2025dexterous,
  title={Dexterous Grasping with Real-World Robotic Reinforcement Learning},
  author={Huang, Dongchi and Zhang, Tianle and Li, Yihang and Zhao, Ling and Li, Jiayi and Fang, Zhirui and Xia, Chunhe and He, Xiaodong},
  journal={arXiv preprint arXiv:2503.04014},
  year={2025}
}

@inproceedings{nair2018overcoming,
  title={Overcoming Exploration in Reinforcement Learning with Demonstrations},
  author={Nair, Ashvin and McGrew, Bob and Andrychowicz, Marcin and Zaremba, Wojciech and Abbeel, Pieter},
  booktitle={2018 IEEE International Conference on Robotics and Automation (ICRA)},
  pages={6292--6299},
  year={2018},
  organization={IEEE}
}

@misc{chen2026borabridgingofflinereinforcement,
      title={BORA: Bridging Offline Reinforcement Learning and Online Residual Adaptation for Real-World Dexterous VLA Models}, 
      author={Zhongxi Chen and Yifan Han and Yanming Shao and Huanming Liu and Congsheng Xu and Xiaoyu Chen and Yao Mu and Wenzhao Lian},
      year={2026},
      eprint={2605.30226},
      archivePrefix={arXiv},
      primaryClass={cs.RO},
      url={https://arxiv.org/abs/2605.30226}, 
}

@inproceedings{wu2025learning,
  title={Learning from imperfect demonstrations with self-supervision for robotic manipulation},
  author={Wu, Kun and Liu, Ning and Zhao, Zhen and Qiu, Di and Li, Jinming and Che, Zhengping and Xu, Zhiyuan and Tang, Jian},
  booktitle={2025 IEEE International Conference on Robotics and Automation (ICRA)},
  pages={16899--16906},
  year={2025},
  organization={IEEE}
}

@article{luo2025precise,
  title={Precise and dexterous robotic manipulation via human-in-the-loop reinforcement learning},
  author={Luo, Jianlan and Xu, Charles and Wu, Jeffrey and Levine, Sergey},
  journal={Science Robotics},
  volume={10},
  number={105},
  pages={eads5033},
  year={2025},
  publisher={American Association for the Advancement of Science}
}

@misc{simeoni2025dinov3,
      title={DINOv3}, 
      author={Oriane Siméoni and Huy V. Vo and Maximilian Seitzer and Federico Baldassarre and Maxime Oquab and Cijo Jose and Vasil Khalidov and Marc Szafraniec and Seungeun Yi and Michaël Ramamonjisoa and Francisco Massa and Daniel Haziza and Luca Wehrstedt and Jianyuan Wang and Timothée Darcet and Théo Moutakanni and Leonel Sentana and Claire Roberts and Andrea Vedaldi and Jamie Tolan and John Brandt and Camille Couprie and Julien Mairal and Hervé Jégou and Patrick Labatut and Piotr Bojanowski},
      year={2025},
      eprint={2508.10104},
      archivePrefix={arXiv},
      primaryClass={cs.CV},
      url={https://arxiv.org/abs/2508.10104}, 
}

@misc{feng2026procvlmlearningproceduregroundedprogress,
      title={ProcVLM: Learning Procedure-Grounded Progress Rewards for Robotic Manipulation}, 
      author={Youhe Feng and Hansen Shi and Haoyang Li and Xinlei Guo and Yang Wang and Chengyang Zhang and Jinkai Zhang and Xiaohan Zhang and Jie Tang and Jing Zhang},
      year={2026},
      eprint={2605.08774},
      archivePrefix={arXiv},
      primaryClass={cs.RO},
      url={https://arxiv.org/abs/2605.08774}, 
}

@misc{mao2026armadvantagerewardmodeling,
      title={ARM: Advantage Reward Modeling for Long-Horizon Manipulation}, 
      author={Yiming Mao and Zixi Yu and Weixin Mao and Yinhao Li and Qirui Hu and Zihan Lan and Minzhao Zhu and Hua Chen},
      year={2026},
      eprint={2604.03037},
      archivePrefix={arXiv},
      primaryClass={cs.RO},
      url={https://arxiv.org/abs/2604.03037}, 
}


%% file: references.bib
@IEEEtranBSTCTL{IEEEauthorLimit,
  CTLuse_forced_etal = {yes},
  CTLmax_names_forced_etal = {6},
  CTLnames_show_etal = {6},
  CTLdash_repeated_names = {no}
}

@misc{shi2026memoryvlaperceptualcognitivememoryvisionlanguageaction,
      title={MemoryVLA: Perceptual-Cognitive Memory in Vision-Language-Action Models for Robotic Manipulation}, 
      author={Hao Shi and Bin Xie and Yingfei Liu and Lin Sun and Fengrong Liu and Tiancai Wang and Erjin Zhou and Haoqiang Fan and Xiangyu Zhang and Gao Huang},
      year={2026},
      eprint={2508.19236},
      archivePrefix={arXiv},
      primaryClass={cs.RO},
      url={https://arxiv.org/abs/2508.19236}, 
}

@misc{zhou2025dinowmworldmodelspretrained,
      title={DINO-WM: World Models on Pre-trained Visual Features enable Zero-shot Planning}, 
      author={Gaoyue Zhou and Hengkai Pan and Yann LeCun and Lerrel Pinto},
      year={2025},
      eprint={2411.04983},
      archivePrefix={arXiv},
      primaryClass={cs.RO},
      url={https://arxiv.org/abs/2411.04983}, 
}

@misc{edstedt2026romav2harderbetter,
      title={RoMa v2: Harder Better Faster Denser Feature Matching}, 
      author={Johan Edstedt and David Nordström and Yushan Zhang and Georg Bökman and Jonathan Astermark and Viktor Larsson and Anders Heyden and Fredrik Kahl and Mårten Wadenbäck and Michael Felsberg},
      year={2026},
      eprint={2511.15706},
      archivePrefix={arXiv},
      primaryClass={cs.CV},
      url={https://arxiv.org/abs/2511.15706}, 
}
